\documentclass[letterpaper, 10 pt, conference]{ieeeconf}  

\IEEEoverridecommandlockouts                              

\usepackage{microtype}
\usepackage{graphicx}
\usepackage{subfigure}
\usepackage{booktabs} 
\usepackage{balance}

\makeatletter
\let\NAT@parse\undefined
\makeatother
\usepackage{hyperref}
\usepackage[numbers,sort&compress]{natbib}

\usepackage{amssymb}
\usepackage{multirow}
\usepackage{amsmath}
\usepackage{algorithm,algpseudocode}
\usepackage{color}
\usepackage{soul}
\usepackage{xspace}
\usepackage{cleveref}

\Crefname{figure}{Figure}{Figures}
\crefname{figure}{Figure}{Figures}
\crefname{table}{Table}{Tables}
\crefformat{table}{Table~#1}
\crefformat{equation}{eqn.(#1)}
\Crefformat{equation}{Eqn.(#1)}
\crefformat{algorithm}{Algorithm~#1}

\newif\ifcomments
\commentsfalse 

\ifcomments
	\newcommand{\dXX}[1]{\color{red}DK: (#1)\color{black}\xspace}  
	\newcommand{\cXX}[1]{\color{blue}CS: (#1)\color{black}\xspace}  
	\newcommand{\XX}[1]{\color{green}JH: (#1)\color{black}\xspace}  
\else
    \newcommand{\dXX}[1]{}  
	\newcommand{\cXX}[1]{}  
	\newcommand{\XX}[1]{}  
\fi

\newcommand{\bea}{\begin{eqnarray}}
\newcommand{\eea}{\end{eqnarray}}
\newcommand{\beas}{\begin{eqnarray*}}
\newcommand{\eeas}{\end{eqnarray*}}
\newcommand{\leftm}{\left[\begin{array}}
\newcommand{\rightm}{\end{array}\right]}
\newcommand{\reals}{\mbox{$\mathbb R$}}

\def\span{{\mbox{\bf span}}}

\newcommand{\mC}{\mathcal{C}}
\newcommand{\mA}{\mathcal{A}}
\newcommand{\mL}{\mathcal{L}}

\newcommand{\mI}{\mathcal{I}}
\newcommand{\mS}{\mathcal{S}}
\newcommand{\mSy}{\mathbb{S}}
\newcommand{\mE}{\mathbb{E}}

\definecolor{commentclr}{RGB}{110, 149, 204}

\newtheorem{lem}{Lemma}

\title{\LARGE \bf
FISAR: Forward Invariant Safe Reinforcement Learning\\with a Deep Neural Network-Based Optimizer
}

\author{Chuangchuang Sun$^{1}$ Dong-Ki Kim$^{1}$ and Jonathan P. How$^{1}$
\thanks{$^{1}$Laboratory for Information \& Decision Systems (LIDS), Massachusetts Institute of Technology, 77 Massachusetts Ave, Cambridge, MA 02139.
{\{\tt\small ccsun1,dkkim93,jhow\}@mit.edu}. This work was support in part by ARL DCIST under Cooperative Agreement Number W911NF-17-2-0181.} %
}

\begin{document}

\maketitle
\thispagestyle{empty}
\pagestyle{empty}

\begin{abstract}
This paper investigates reinforcement learning with constraints, which are indispensable in safety-critical environments. To drive the constraint violation to decrease monotonically, we take the constraints as Lyapunov functions and impose new linear constraints on the policy parameters' updating dynamics. As a result, the original safety set can be forward-invariant. However, because the new guaranteed-feasible constraints are imposed on the updating dynamics instead of the original policy parameters, classic optimization algorithms are no longer applicable. To address this, we propose to learn a generic deep neural network (DNN)-based optimizer to optimize the objective while satisfying the linear constraints. The constraint-satisfaction is achieved via projection onto a polytope formulated by multiple linear inequality constraints, which can be solved analytically with our newly designed metric. To the best of our knowledge, this is the \textit{first} DNN-based optimizer for constrained optimization with the forward invariance guarantee. We show that our optimizer trains a policy to decrease the constraint violation and maximize the cumulative reward monotonically. Results on numerical constrained optimization and obstacle-avoidance navigation validate the theoretical findings.

\end{abstract}

\section{Introduction}\label{sec:introduction}
Reinforcement learning (RL) has achieved remarkable success in robotics~\cite{schulmanetal-16-gae,levine16end2end,gu17manipulationrl}.
In general, an RL agent is free to explore the entire state-action space and improves its performance via trial and error~\cite{sutton2018reinforcement}. 
However, there are many safety-critical scenarios where an agent cannot explore certain regions. 
For example, a self-driving vehicle must stay on the road and avoid collisions with other cars and pedestrians. 
An industrial robot also should not damage the safety of the workers. 
Another example is a medical robot, which should not endanger a patient's safety. 
Therefore, an effective agent should satisfy certain safety constraints during its exploration, and failure to do so can result in undesirable outcomes. 

The safe exploration problem can be represented by the constrained Markov decision process (CMDP)~\cite{altman1999constrained}.
Existing optimization techniques to solve CMDP include the vanilla Lagrangian method~\cite{chow2017risk}, which solves a minimax problem by alternating between primal policy and dual variables. Further, a PID Lagrangian method in~\cite{stooke2020responsive} addresses the oscillations and overshoot in learning dynamics that lead to constraint violation. However, these methods show difficulties when solving a minimax problem with non-convexity (e.g., non-linear function approximations).
Another approach solves CMDP as non-convex optimization directly via successive convexification of the objective and constraints~\cite{achiam2017constrained, yu2019convergent}. 
However, the convexification methods also have several drawbacks: 1) there is a lack of understanding of how the constraint is driven to be feasible (e.g., at what rate does the constraint violation converge to zero?), 2) the convexified subproblem can often encounter infeasibility, requiring a heuristic to recover from infeasibility, and 3) it needs to solve convex programming with linear/quadratic objective and quadratic constraints at every iteration, which is inefficient. 

\begin{figure}[t]
\centering
\includegraphics[scale=0.45]{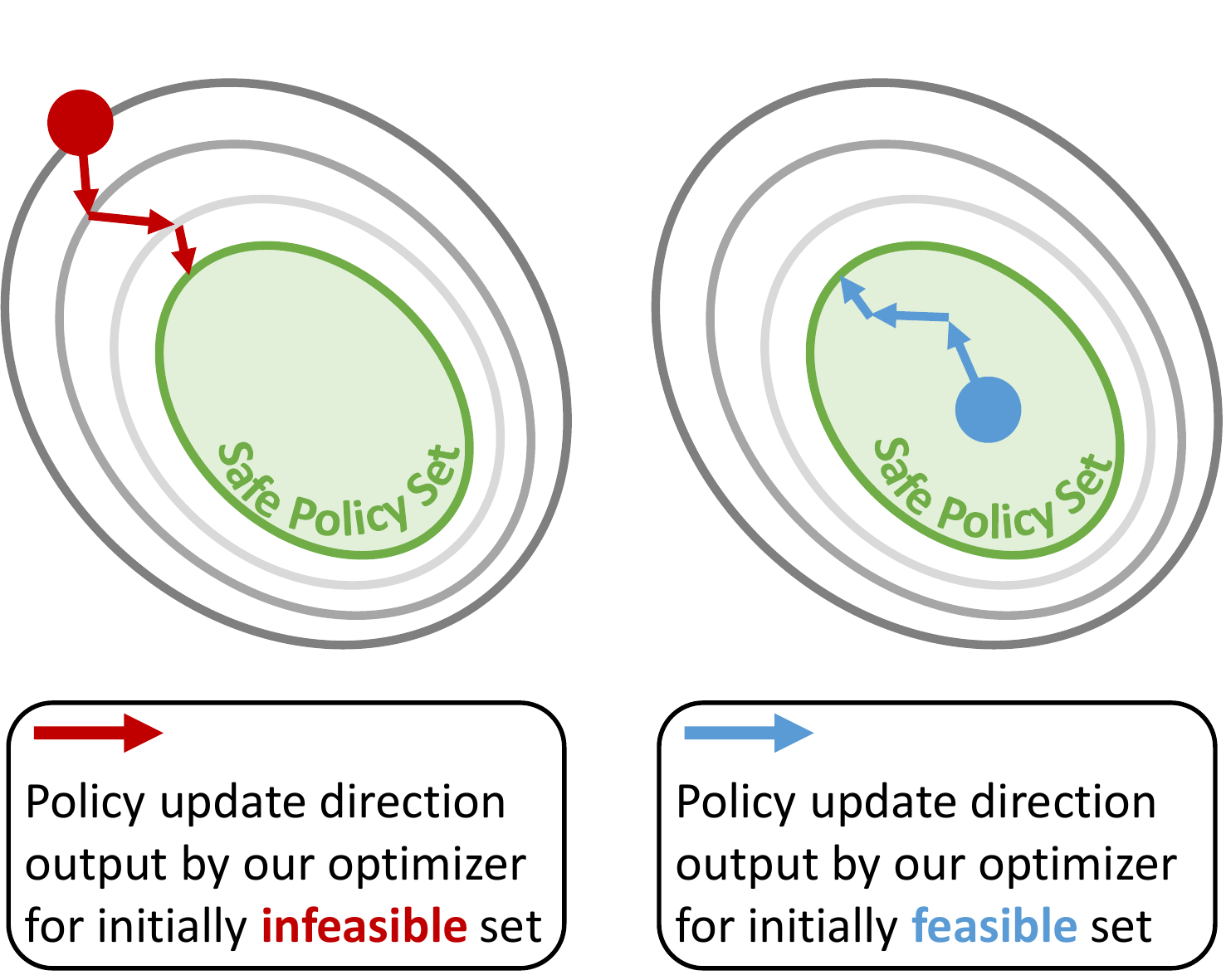}
\vspace{-.2cm}
\caption{Illustration of the forward invariance in the policy space. The contour is for the constraint function, where the darker color denotes a larger violation. Simultaneously optimizing the objective (not shown), our optimizer can guarantee the \textit{forward invariance}: the constraint can converge to satisfaction asymptotically if the policy initialization is infeasible, and the trajectory will stay inside the feasible set if the policy initially starts there.}
\label{Fig:forward-invariance}
\end{figure}

In this paper, we introduce a new learning-based framework to address the aforementioned limitations in solving CDMP. 
Specifically, we propose to take safety constraints as Lyapunov functions to drive the constraint violation monotonically decrease and impose new constraints on the updating policy dynamics. 
We note that such new constraints, which are linear inequalities and guaranteed to be feasible, can guarantee the \textit{forward invariance}: the constraint violation can converge asymptotically if the policy initialization is infeasible, and the trajectory will stay inside the feasible set if the policy initially starts there (see~\cref{Fig:forward-invariance}).
However, with the new constraints imposed on the policy update dynamics, it is difficult to design such updating rules to optimize the objective while simultaneously satisfying the constraints. 
Methods like projected gradient descent~\cite{nesterov2013introductory} are not applicable here because the constraints are on the updating dynamics instead of on the primal variables. 
Therefore, we propose to learn an optimizer parameterized by a deep neural network, where the constraint-satisfaction is guaranteed by projecting the optimizer output onto those linear inequality constraints. 
While generic projection onto polytopes formulated by multiple linear inequalities cannot be solved in closed form, we design a proper metric for the projection such that it can be solved analytically. 

\textbf{Contribution.}
In summary, our contributions are twofold. 
First, We propose a model-free framework to learn a deep neural network-based optimizer to solve a safe RL problem formulated as a CMDP with guaranteed feasibility without solving a constrained optimization problem iteratively, unlike the algorithms based on successive convexification~\cite{achiam2017constrained, yu2019convergent}. 
To the best of our knowledge, this is the \textit{first} generic DNN-based optimizer for constrained optimization and can be applied beyond the safe learning context. 
Second, the resulting updating dynamic of the policy parameters implies forward-invariance of the safety set. Hence, our method theoretically guarantees that the constraint violation will converge asymptotically, which has not been established yet among existing safe reinforcement learning works.

\section{Related Works}
\textbf{Safe reinforcement learning.} 
Algorithms from a control-theoretic perspective mainly fall into the category of Lyapunov methods. 
For tabular settings, Lyapunov functions are constructed in \cite{chow2018lyapunov} to guarantee global safety during training via a set of local linear constraints. 
Another work obtains high-performance control policies with provable stability certificates using the Lyapunov stability verification~\cite{berkenkamp2017safe}.
Recently, \cite{richards2018lyapunov} constructs a neural network Lyapunov function and trains the network to the shape of the largest safe region in the state space.

On the other hand, the control barrier function~\cite{ames2016control} provides a venue to calibrate the potentially unsafe control input to the safety set. 
For example, \cite{cheng2019end} introduces an end-to-end trainable safe RL method, which compensates the control input from the model-free RL via model-based control barrier function. 
To avoid solving an optimization problem while guaranteeing safety, the vertex network \cite{zheng2020safe} formulates a polytope safety set as a convex combination of its vertices. 
In~\cite{choi2020reinforcement}, an input-output linearization controller is generated via a control barrier function and control Lyapunov function based quadratic program with the model uncertainty learned by reinforcement learning.

There are also approaches to solve a safe RL problem with temporal logic specifications~\cite{alshiekh2017safe,fulton2018safe} and curriculum learning~\cite{turchetta2020safe}.
See~\cite{garcia2015comprehensive} for in-depth surveys about safe RL.

Compared to these approaches, our method is based on the model-free policy gradient reinforcement learning, so neither the transition dynamics nor the cost function is explicitly needed. 
Additionally, our approach guarantees forward invariance, so the policy will be updated to be only safer.

\textbf{DNN-based optimizer.} In contrast to hand-designed optimization algorithms, \cite{andrychowicz2016learning} proposes to cast the design of a gradient-based optimization algorithm as a learning algorithm. 
This work is then further extended to learn a gradient-free optimizer in~\cite{chen2017learning}. 
Recently, \cite{li2017meta} introduces the Meta-SGD, which can initialize and adapt to any differentiable learner in just one step.
This approach shows a highly competitive performance for few-shot learning settings. 
To improve the scalability and generalization of DNN-based optimizers, \cite{wichrowska2017learned} develops a hierarchical recurrent neural network architecture that uses a learned gradient descent optimizer. 
For more information about deep neural network-based optimizers, we refer to the survey~\cite{hospedales2020meta}. 
However, these previous works are designed for unconstrained optimization. 
In this paper, we extend these approaches and develop a DNN-based optimizer for constrained optimization.

\section{Preliminary}\label{sec:preliminary}
\subsection{Markov decision process}\label{sec:mdp}
The Markov decision process (MDP) is a tuple $\langle \mS,\mA,\mathcal{T},\mathcal{R}, \gamma, P_0 \rangle$, where $\mS$ is the set of the agent state in the environment, $\mA$ is the set of agent actions, $\mathcal{T}:\mS\times \mA\times \mS\to[0,1]$ is the transition function, $\mathcal{R}$ denotes the reward function, $\gamma\in[0,1]$ is the discount factor and $P_0:\mS\to [0,1]$ is the initial state distribution.
A policy $\pi: \mS\to P (\mA)$ is a mapping from the state space to probability over actions. $\pi_\theta(a|s)$ denotes the probability of taking action $a$ under state $s$ following a policy parameterized by $\theta$. 
The objective is to maximize the cumulative reward: 
\begin{equation}\label{eq:objective}
J(\theta) = \mE_{\tau\sim p_\theta(\tau)}[\sum_t \gamma^t \mathcal{R}(s_t, a_t)],
\end{equation}
where $J(\theta):\reals^n\to\reals$ and $\tau$ are trajectories sampled under $\pi_\theta(a|s)$.
To optimize the policy that maximizes \eqref{eq:objective}, the policy gradient with respect to $\theta$ can be computed as~\cite{sutton2000policy}:
$
\nabla_\theta J(\theta) = \mE_{\tau\sim \pi_\theta(\tau)}[\nabla_\theta \log\pi_\theta(\tau) G(\tau)],
$
where $G(\tau) = \sum_{t} \gamma^t \mathcal{R}(s_t, a_t)$~\cite{sutton2018reinforcement}.

\subsection{Constrained Markov decision process}
The constrained Markov decision process (CMDP) is defined as a tuple $\langle \mS,\mA,\mathcal{T},\mathcal{R}, c, \gamma, P_0 \rangle$, where $c:\mS\times \mA\times \mS\to\reals$ is the cost function and the other variables are identical to those in the MDP definition (see~\Cref{sec:mdp})~\cite{altman1999constrained}. The goal in CMDP is to maximize the cumulative reward {while} satisfying the constraints on the cumulative cost:
\vspace{-.2cm}
\begin{eqnarray}\label{eq:cmdp}
&&\hspace{-1cm} \max_\theta J(\theta) = \mE_{\tau\sim \pi_\theta(\tau)}\Big[\sum_t \gamma^t r(s_t, a_t)\Big],\\
&&\hspace{-1cm} s.t.\ C_i(\theta) = \mE_{\tau\sim \pi_\theta(\tau)}\Big[\sum_t \gamma^t c_i(s_t, a_t)\Big]-\Bar{C}_i\le0, i\in\mathcal{I}\nonumber
\end{eqnarray}
where $\theta\in \reals^n$ is the policy parameters, $C_i(\theta):\reals^n\to\reals$, $\mathcal{I}$ is the constraint set, and $\Bar{C}_i$ is the maximum acceptable violation of $C_i(\theta)$. 
In a later context, we use $J$ and $C_i$ as short-hand versions of $J(\theta)$ and $ C_i(\theta)$, respectively, for clarity. While the discount factor for the cost can be different from that for the reward, we use the same for notational simplicity.

Instead of imposing safety constraints on the cumulative cost (see \eqref{eq:cmdp}), there is another option of imposing them on individual state-action pairs \cite{cheng2019end,zheng2020safe}:
\begin{eqnarray}\label{eq:cmdp_variant}
&&\hspace{-1cm} \max_\theta J(\theta) = \mE_{\tau\sim \pi_\theta(\tau)}\Big[\sum_t \gamma^t r(s_t, a_t)\Big],\\
&&\hspace{-1cm} s.t.\ c_i(s_t, a_t) \in \mC_i, \forall i\in\mathcal{I}, \forall t \le T_{\max},\nonumber
\end{eqnarray}
where $t\in \mathbb{N}$ and $T_{\max}\in \mathbb{N}$ denotes the horizon of the MDP. 
We note that \eqref{eq:cmdp}, the problem formulation considered in this work, is more general than \eqref{eq:cmdp_variant} in two aspects. 
First, the constraint on each individual state-action pair can be transformed into the form of cumulative cost via setting binary cost function followed by summation with $\gamma=1$ in a finite horizon MDP. 
That is to say, the constraints in~\eqref{eq:cmdp_variant} can be re-written equivalently in the form of the constraints in~\eqref{eq:cmdp} as $\sum_t \textbf{1}_{\mC_i}(c_i(s_t, a_t)) \le 0$, where $\textbf{1}_\mC(x)$ is an indicator function such that $\textbf{1}_\mC(x) = 0$ if $x\in\mC$ and $\textbf{1}_\mC(x) = 1$ otherwise. 
Second, in scenarios where the agent can afford a certain amount of violations (i.e., $\Bar{C}_i$ in \eqref{eq:cmdp}) throughout an episode, it is infeasible to allocate it to individual time instances. 
An example scenario is a video game, where a player can stand some given amount of attacks in the lifespan before losing the game.

\subsection{Control barrier functions}
Consider the following non-linear control-affine system:
\bea\label{eq:system}
\dot{x} = f(x) + g(x)u,
\eea
where $f$ and $g$ are locally Lipschitz, $x \in D \subset \reals^n$ is the state and $u \in U \subset \reals^m$ is the set of admissible inputs. The safety set is defined as $\mC = \{x \in D \subset \reals^n| h(x) \le 0\}$ with $\mC\subset D$. Then $h$ is a control barrier function (CBF)~\cite{ames2016control} if there exists an extended class-$\kappa_{\infty}$ function $\alpha$ such that for the control system \eqref{eq:system}:
\bea\label{eq:cbf}
\sup_{u\in U}(L_fh(x) + L_gh(x)u) \le -\alpha(h(x)), \forall x \in D,
\eea
where $L_fh(x) = \big( \frac{\partial h(x)}{\partial x} \big)^Tf(x)$ is the Lie derivative.

\section{Approach}\label{sec:approach}
\subsection{Forward-invariant constraints on updating dynamics}
The key to solving~\eqref{eq:cmdp} is how to deal with the constraints. 
Existing methods~\cite{chow2017risk,achiam2017constrained} often encounter oscillations and overshoot~\cite{stooke2020responsive} in learning dynamics that can result in noisy constraint violations. 
In this paper, we aim to address this issue and build a new mechanism that drives the constraint violation to converge asymptotically if the initialization is infeasible. Otherwise, the trajectory will stay inside the feasible set (i.e., forward invariance). 
To accomplish our goal, we start by building a Lyapunov-like condition:
\bea\label{eq:lyaponuv}
\frac{\partial C_i}{\partial \theta}\dot{\theta} \le -\alpha(C_i(\theta)),i\in\mI,
\eea
where $\dot{\theta}$ is the updating dynamics of $\theta$ and $\alpha(\bullet)$ is an extended class-$\kappa$ function. Note that such Lyapunov functions are directly taken from the constraint functions so that this process needs no extra effort. A special case of the class-$\kappa$ function is a scalar linear function with positive slope and zero intercept. With discretization, the updating rule becomes: 
\bea\label{eq:update}
\theta_{k+1} = \theta_k + \beta \dot{\theta}_k,
\eea
where $\beta > 0$ denotes the learning rate.
Note that, with sufficiently small $\beta$, the continuous dynamics can be approximated with a given accuracy.
\Cref{lem:invariance} characterize how \eqref{eq:lyaponuv} will make the safety set $\mC = \{\theta|C_i\le0,\forall i\in\mI\}$ forward invariant. For notational simplicity, the statement is on one constraint $\mC_i$ with $\mC=\cap_{i\in \mI} \mC_i$ and $\mC_i = \{\theta|C_i\le0,i\in\mI\}$.
This simplification does not lose any generality because the joint forward-invariance of multiple sets will naturally lead to the forward-invariance of their intersection set.

\begin{lem}\label{lem:invariance}
Consider a continuously differentiable set $\mC_i = \{\theta|C_i\le0,i\in\mI\}$ with $C_i$ defined on $\mathcal{D}$. Then $\mC_i$ is forward invariant, if $\mathcal{D}$ is a superset of $\mC_i$ (i.e., $\mC \subseteq \mathcal{D} \subset \reals^n$), and \eqref{eq:lyaponuv} is satisfied.
\end{lem}
\emph{Proof: } Define $\partial \mC_i = \{\theta|C_i(\theta) = 0, i\in\mI\}$ as the boundary of $\mC_i$. As a result, for $\theta\in \partial\mC_i$, $\frac{\partial C_i}{\partial \theta}\dot{\theta} \le -\alpha(C(\theta)) = 0$. 
Then, according to the Nagumo’s theorem~\cite{blanchini2008set, blanchini1999set}, the set $\mC_i$ is forward invariant.$\blacksquare$

Here we provide intuition behind~\eqref{eq:lyaponuv}. 
Using the chain rule: $\frac{\partial C_i(\theta(t))}{\partial t} = \frac{\partial C_i}{\partial \theta}\dot{\theta} = -C_i(\theta(t))$. 
Then, the solution to this partial differential equation is $C_i(t) = ce^{-t}$. With $c>0$, it means that the initialization is infeasible (i.e., $C_i(0) > 0$), and thus $C(t)$ will converge to $0$ (i.e., the boundary of $C_i$) asymptotically. It is similar with a feasible initialization (i.e., $c\le0$). 
It is worth noting that with $|\mI| \le n$, i.e., the number of constraints is smaller than that of the policy parameters, \eqref{eq:lyaponuv} is guaranteed to be feasible. 
This saves the trouble of recovering from infeasibility in a heuristic manner, which is usually the case for the previous approaches~\cite{achiam2017constrained, yu2019convergent}.


While \eqref{eq:lyaponuv} in our proposed method looks similar to \eqref{eq:cbf} in CBF, our method is substantially different from those exploiting CBF~\cite{cheng2019end, zheng2020safe}. 
First, CBF-based methods require the system dynamics in \eqref{eq:system} while our method is model-free, not requiring transition dynamics and cost function $c_i(s_t, a_t)$ in \eqref{eq:cmdp} (in parallel to $h(x)$ in \eqref{eq:cbf}). 
Second, it is more significant that \eqref{eq:cbf} and \eqref{eq:lyaponuv} represent different meanings. 
On one hand, the former represents the constraint on the control input $u_t$, given a state $x_t$ at a certain time instance $t$, while in the latter, $C_i$ is evaluated on multiple time instances (e.g., one episode). 
Due to this, considering multiple time instances can help make a globally optimal decision while one-step compensation can be short-sighted. On the other hand, if we further replace $u_t$ by $u_{\theta}(x_t)$, a policy parameterized by $\theta$, \eqref{eq:cbf} becomes a \textit{non-linear} constraint on policy parameter $\theta$ at time instance $t$, while \eqref{eq:lyaponuv} is a constraint imposed on $\dot{\theta}$ instead of $\theta$. 
Such constraint on the updating dynamics $\dot{\theta}$ can result in \textit{forward invariance directly} in the policy space ($\theta$ therein). 
By contrast, the forward-invariance of CBF is in the state space ($x$ therein), and thus it still requires to solve an optimization problem to generate a control input~\cite{cheng2019end} at each time instance, which can be computationally inefficient.



\subsection{Learning a deep neural network-based optimizer}
So far, we have converted the constraint on $\theta$ in \eqref{eq:cmdp} to that on $\dot{\theta}$ in \eqref{eq:lyaponuv}, which formulates the new set
\bea\label{eq:cbf_policy}
\mC_{i,\dot{\theta}} = \Big\{\dot{\theta}|\frac{\partial C_i}{\partial \theta}\dot{\theta} \le -\alpha(C_i(\theta)),i\in\mI\Big\},
\eea
and $\mC_{\dot{\theta}} = \cap_{i\in\mI}\mC_{i,\dot{\theta}}$. However, it is unclear how to design an optimization algorithm that minimizes the objective in \eqref{eq:cmdp} while satisfying \eqref{eq:lyaponuv}. 
Note that the typical constrained optimization algorithms, such as projected gradient descent (PGD), are no longer applicable because the constraints are not on the primal variables anymore. 
Specifically, similar to the PGD mechanism, $\theta$ can be updated in the following way:
\bea\label{eq:pgd}
\theta_{k+1} = \theta_k + \beta \text{proj}_{\mC_{\dot{\theta}}}(\nabla_\theta J(\theta_k)),
\eea
where $\text{proj}_{\mC_{\dot{\theta}}}(\bullet)$ is the projection operator onto the set ${\mC_{\dot{\theta}}}$. However, this can be problematic as it is ambiguous whether
$\text{proj}_{\mC_{\dot{\theta}}}(\nabla_\theta J(\theta_k))$ is still an ascent direction.
Consequently, standard optimization algorithms (e.g., stochastic gradient descent (SGD), ADAM~\cite{kingma2014adam}) with \eqref{eq:pgd}, will fail to optimize the objective while satisfying the constraints as we will show in the result section. 
Thus, we propose to learn a DNN-based optimizer.

\begin{figure}[t]
\centering
\includegraphics[scale=0.55]{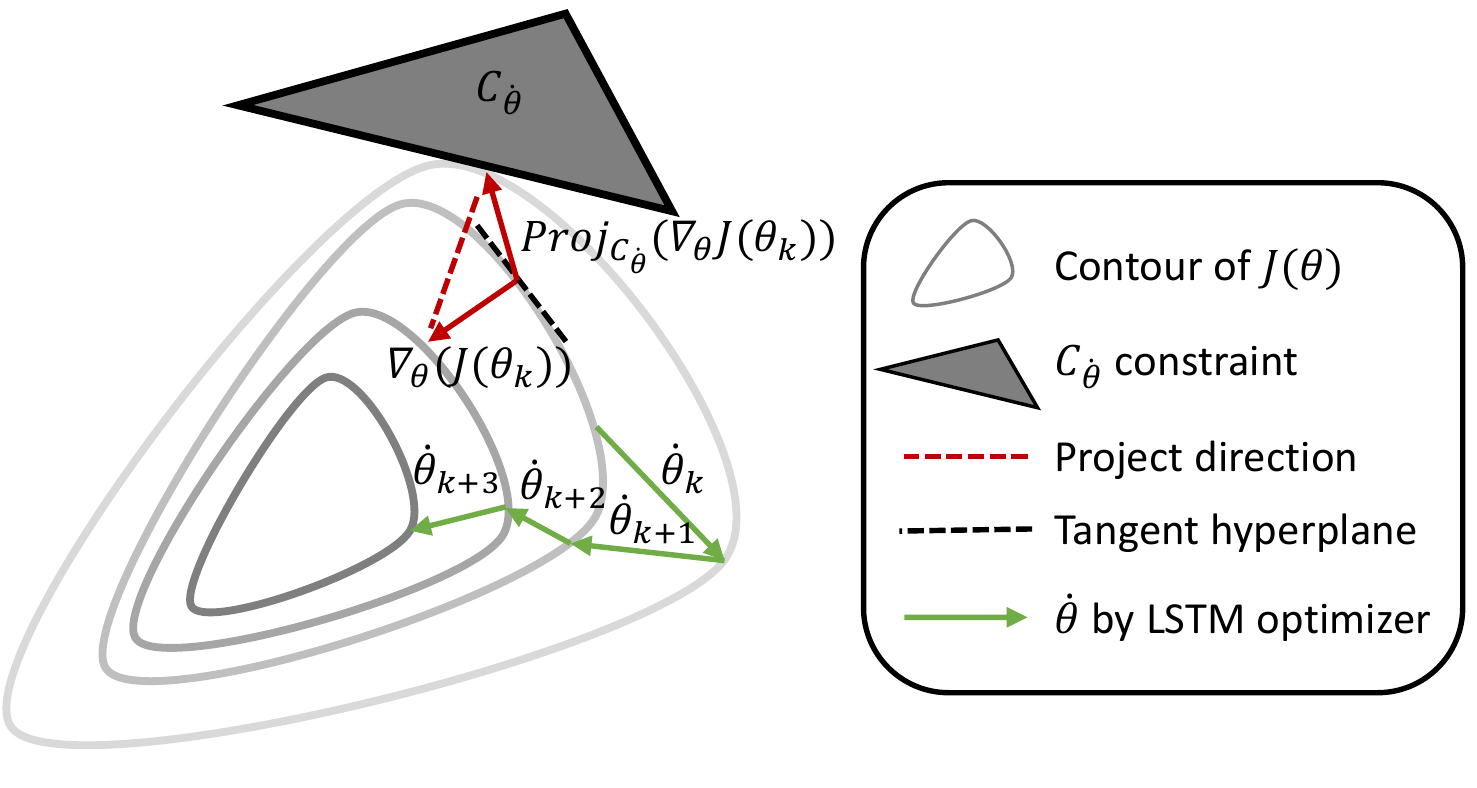}
\vspace{-.2cm}
\caption{Comparison between the projected gradient descent in \eqref{eq:pgd} and LSTM-based optimizer in \eqref{eq:optimizer}. Considering the maximization problem of $J(\theta)$, so $\nabla_\theta J(\theta)$ is the ascent direction (towards the darker contour line). The one-step projected can lead to an undesired descent direction and the performance throughout the iterations cannot be guaranteed. On the contrary, LSTM optimizer consider the whole optimization iteration span and can achieve an optimal objective value at the last iteration even with some intermediate objective descents.}
\label{Fig:optimizer_comparison}%
\end{figure}

Following the work by \cite{andrychowicz2016learning}, which learns an optimizer for unconstrained optimization problems, we extend it to the domain of constraint optimization. 
Our optimizer is parameterized by a long short-term memory (LSTM, \cite{hochreiter1997long}) $m_{\phi}$ with $\phi$ as the parameters for the LSTM network $m$. 
We note that the recurrence nature of LSTM allows to learn
dynamic update rules by fully exploiting historical gradient information, similar to the momentum-based optimization techniques~\cite{nesterov27method,kingma2014adam}. 
Similar to \cite{andrychowicz2016learning}, the updating rule becomes:
\bea\label{eq:optimizer}
&&\theta_{k+1} = \theta_k + \beta \dot{\theta}_k \nonumber\\
&&\dot{\theta}_k = \text{proj}_{\mC_{\dot{\theta}}}(\dot{\theta}_k^-) \\
&&\begin{bmatrix}
\dot{\theta}_k^-\\
h_{k+1}
\end{bmatrix}= m_{\phi}(\nabla_{\theta}(J(\theta_k)), h_k) \nonumber,
\eea
where $h_k$ is the hidden state for $m_{\phi}$. 
The loss to train the optimizer parameter $\phi$ is defined as:
\bea\label{eq:loss}
\mL(\phi) = -\mE_f\Bigg[\sum_{k=1}^{T_{\phi}} w_k J(\theta_k)\Bigg],
\eea
where $T_{\phi}$ is the span of the LSTM sequence and $w_k>0$ is the weight coefficient. Given this loss function, $m_\phi$ aims to generate the updating direction of $\theta$ in the whole horizon $k=1\ldots T_\phi$ such that the final $J(\theta_{T_\phi})$ is optimal.
The main difference between ours and \cite{andrychowicz2016learning} is the projection step (i.e., the second line in \eqref{eq:optimizer}). 
As a result, it can be understood that the end-to-end training minimizes the loss in \eqref{eq:loss}
while the constraint-satisfaction is guaranteed by the projection.

Here we take a further qualitative analysis on the difference between the updating rules in \eqref{eq:pgd} and \eqref{eq:optimizer} and validate the advantage of the latter. 
During the iterations of maximizing $J(\theta)$, one-step projected gradient $\text{proj}_{\mC_{\dot{\theta}}}(\nabla_\theta J(\theta_k))$ can result in a descent direction (i.e., the other side of the tangent hyperplane in \cref{Fig:optimizer_comparison}) and is difficult to guarantee performance through the iterations. 
By contrast, $\dot{\theta}$, output from the LSTM optimizer, will take the whole optimization trajectory into consideration (see the loss function \eqref{eq:loss}) to eventually maximize $J(\theta_{T_\phi})$, the objective function value at the last step, even some intermediate steps can have {a few} objective descents as well (e.g., $\dot{\theta}_k$ in \cref{Fig:optimizer_comparison}). 

\begin{figure*}[t]
\centering
\includegraphics[width=0.6\linewidth,page=2]{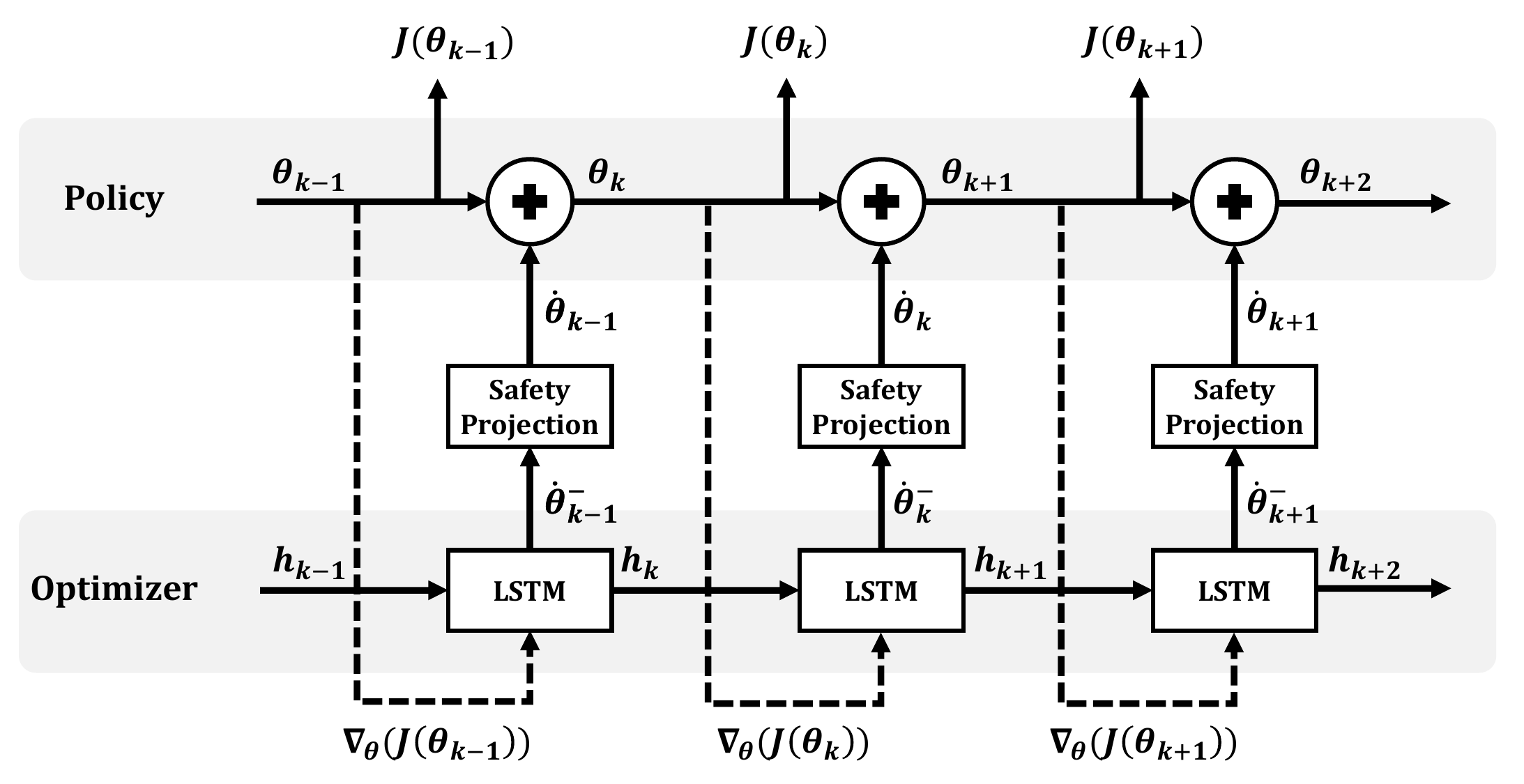}
\vspace{-.1cm}
\caption{Computational graph used for computing the gradient of the neural network-based optimizer. 
The policy (top) is trained based on $\dot{\theta}_k$, the update direction output by the optimizer (bottom) followed by safety projection (center), which takes as input the gradient information $\nabla_{\theta}J(\theta_k)$.
The figure is modified from \cite{andrychowicz2016learning} by adding the safety projection module. Note that gradients are allowed to flow along the solid edges in the graph, not the dashed ones, under the assumption that the gradients of the policy parameters do not depend on the LSTM optimizer parameters. This helps avoid calculating second derivatives, which can be computationally expensive.}
\label{fig:meta-optimizer}
\end{figure*}


\begin{algorithm}[t]
\caption{FISAR: Forward Invariant Safe Reinforcement Learning}
\label{alg:FISAR}
\small
\begin{algorithmic}[1]
\State \textbf{Require:} class $\kappa$ function $\alpha$ in \eqref{eq:cbf_policy}, learning rate $\beta$ in \eqref{eq:optimizer}, weight coefficients $w_k$ in \eqref{eq:loss} and LSTM sequence span length $T_{\phi}$ in \eqref{eq:loss}.
\State  Randomly initialize LSTM optimizer parameter $m_{\phi}$ 
\While {LSTM optimizer parameters not convergent}
    \For {$k=1...T_{\phi}$}
        \State Randomly initialize policy parameter $\theta$
        \State Sample trajectories $\tau$ under policy $\pi_{\theta}(a|s)$
        \State Compute $J(\theta_k)$ via \eqref{eq:objective}
        \State Compute $\nabla_{\theta}(J(\theta_k))$ via \[\nabla_{\theta}(J(\theta_k)) = \mathbb{E}_{\tau\sim \pi_\theta(\tau)}[\nabla_\theta \log \pi_\theta(\tau) r(\tau)] \]
        \State Update $\theta$ via \eqref{eq:optimizer}
    \EndFor
    \State Compute the loss function $\mL(\phi)$ via \eqref{eq:loss}
    \State Update $\phi$: $\phi \leftarrow \phi - \nabla_{\phi}\mL(\phi)$
\EndWhile
\end{algorithmic}
\end{algorithm}

\subsection{Solving projection onto general polytope analytically}
Even $\mC_{\dot{\theta}}$ is a polytope formulated by linear inequalities, projection onto $\mC_{\dot{\theta}}$ is still non-trivial and requires an iterative solver such as in \cite{achiam2017constrained}, except that there is only one inequality constraint (i.e., $|\mI| = 1$). 
Two alternative methods are proposed in \cite{dalal2018safe}: one is to find the single active constraint to transform into a single-constraint case and the other is to take the constraints as a penalty. 
However, the former is troublesome and possibly inefficient and the latter will sacrifice the strict satisfaction of the constraint.

Hence, we propose to solve the projection onto the polytope formulated by multiple linear inequalities in a closed form. 
We first explain the generic projection problem onto a polytope:
\bea\label{eq:qp}
&&\min_x \frac{1}{2}(x-x_0)^TQ(x-x_0), \nonumber\\ 
&&s.t.\ \  Ax \le b,
\eea
where $x_0\in\reals^n$, $A\in\reals^{m\times n}$ is of full row rank and $Q\in\mSy^n$ is positive definite.
Then the dual problem of \eqref{eq:qp} is 
\bea\label{eq:qp_dual}
\min_{\lambda\ge 0} &\frac{1}{2}\lambda^TAQ^{-1}A^T\lambda + \lambda^T(b-Ax_0) 
\eea
The dual problem in \eqref{eq:qp_dual} generally cannot be solved analytically as $AQ^{-1}A^T$ is positive definite but not diagonal. 
Though $Q$ is usually set as the identity matrix, it is not necessary other than that $Q$ should be positive definite. 
As a result, we design $Q$ such that $AQ^{-1}A^T$ is diagonal by solving:
\bea\label{eq:qp_Q}
Q^{-1} &=& \arg\min_H \frac{1}{2} \|H-\delta I\|, \nonumber\\ 
&& s.t.\ \  AHA^T = I,
\eea
where $\delta >0$. As a result, we obtain 
$Q^{-1} = \delta I + A^T(AA^T)^{-1}(I-\delta AA^T)(AA^T)^{-1}A$. Then \eqref{eq:qp_dual} can be solved in closed form as:
\bea\label{eq:lambda}
\lambda &=& \max(0, Ax_0-b), \nonumber\\
x &=& x_0 - Q^{-1}A^T\lambda.
\eea
The schematics of the LSTM-based optimizer is presented in \cref{fig:meta-optimizer} and the algorithm FISAR (\textbf{F}orward \textbf{I}nvariant \textbf{SA}fe \textbf{R}einforcement learning) is summarized in \cref{alg:FISAR}.


\begin{figure}[t]
\centering
\includegraphics[width=0.49\textwidth]{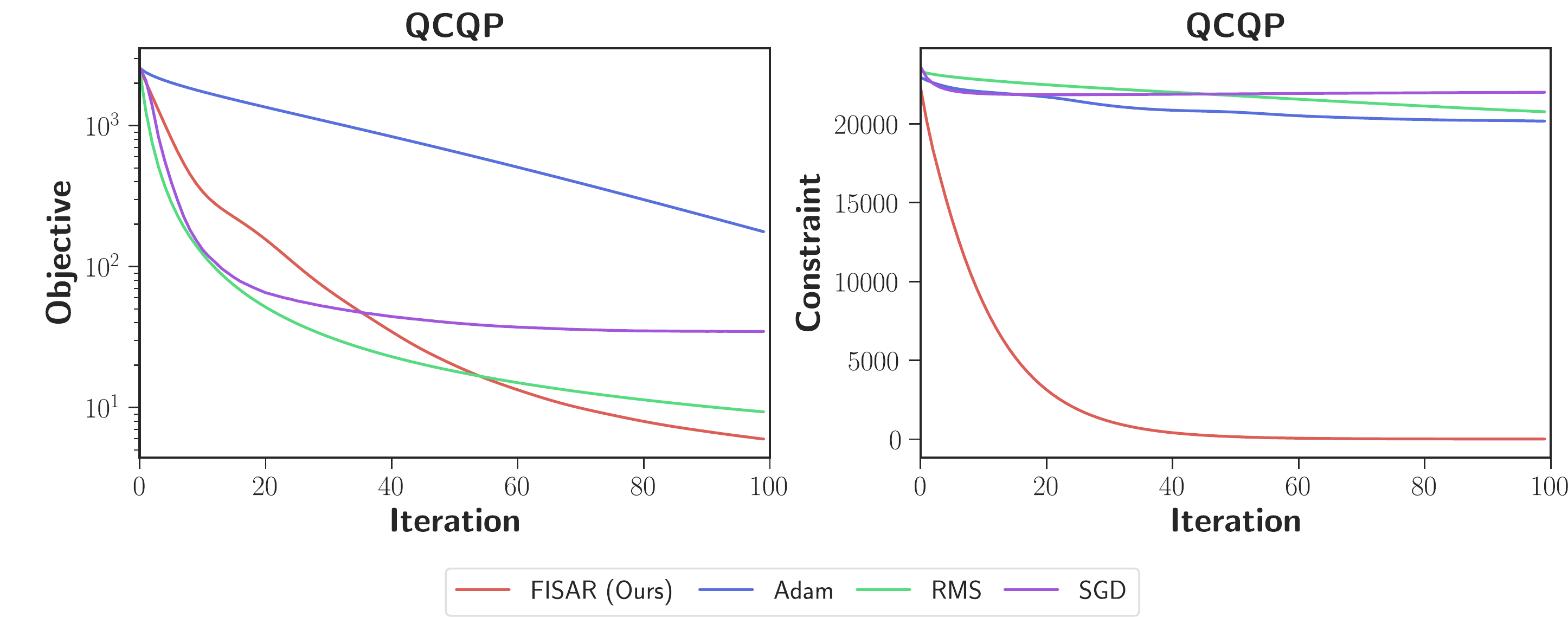}
\vspace{-.6cm}
\caption{The trajectory of the objective (left) and constraint (right) of the deterministic optimization problem \eqref{eq:qcqp} under the learned LSTM-optimizer and three baselines for unconstrained optimization.
The results show that the constraint violation converges to zero asymptotically while our objective is comparable to those achieved by the unconstrained solvers.}%
\label{Fig:qcqp}%
\end{figure}

\section{Experiments}
As our LSTM-parameterized optimizer applies to general constrained optimization problems, it is first tested on a non-linear numerical optimization problem. 
Then, we evaluate our safe RL framework in an obstacle-avoidance navigation environment.


\subsection{Quadratically constrained quadratic programming}
We first apply the learned LSTM-optimizer on the following quadratically constrained quadratic programming (QCQP), which has various applications including signal processing~\cite{huang2014randomized}, graph theory~\cite{goemans1995improved}, and optimal control~\cite{sun2019iterative}. Specifically, the objective and constraints in this domain are defined as:
\bea\label{eq:qcqp}
&&\min_x \|Wx-y\|_2^2, \nonumber\\ 
&&s.t. \ \ (x-x_0)^TM(x-x_0) \le r,
\eea
where $W,M\in\mSy^n$, $x_0,x,y\in\reals^n$ and $r\in\reals$. $M$ here is not necessarily positive semi-definite and thus can bring non-convexity. 

We solve QCQP using our LSTM-optimizer as well as three unconstrained baselines (Adam, RMS, and SGD) to show the scale of the objective. Given the results in \Cref{Fig:qcqp}, in this deterministic setting, the constraint violation is driven to satisfaction asymptotically, while our objective is comparable to that from the unconstrained solvers.

\subsection{Obstacle-avoidance navigation}
\begin{figure}[t]
\centering
\includegraphics[scale=0.27]{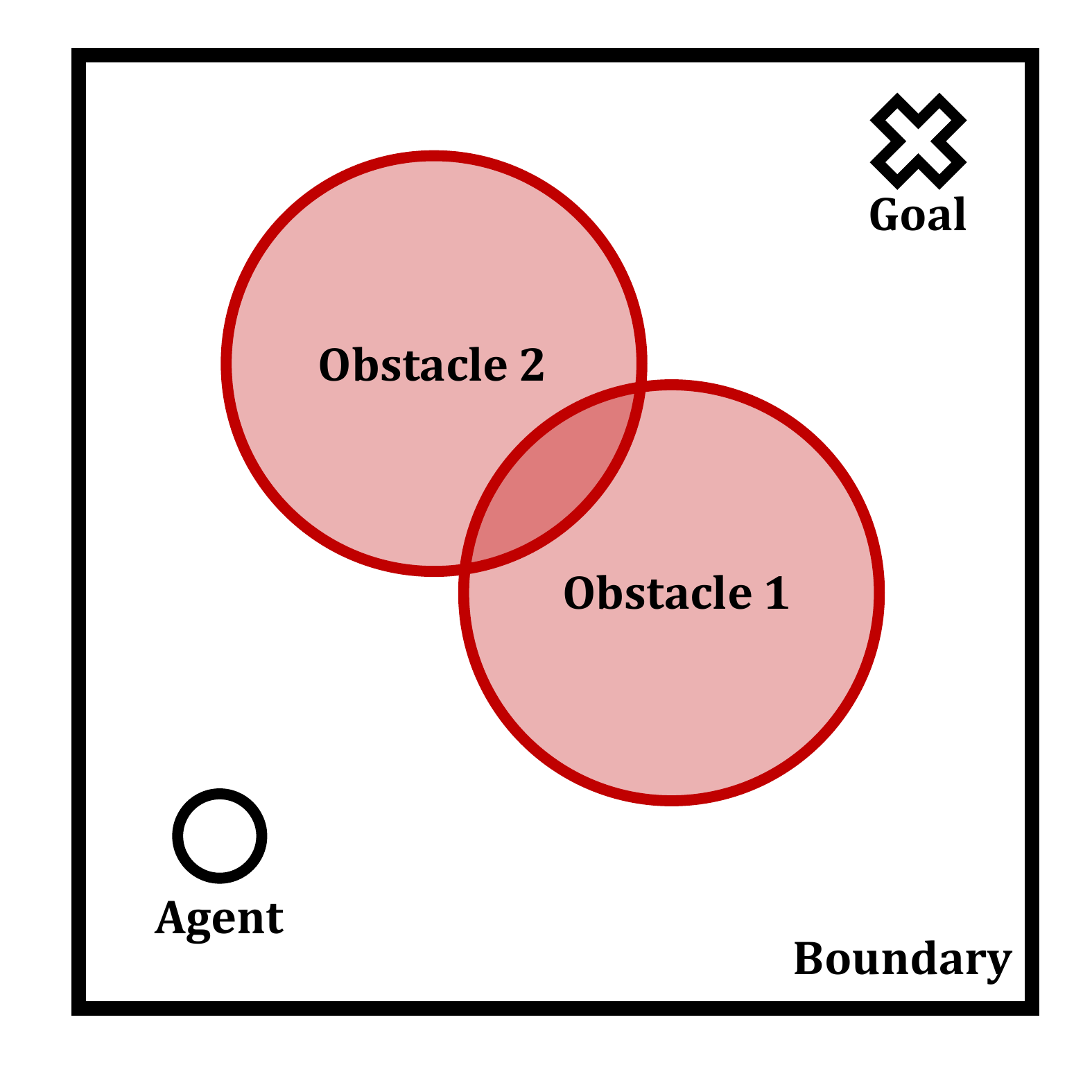}
\vspace{-.cm}
\caption{Illustration of obstacle avoidance navigation environment. The objective is to reach the goal while avoiding the two obstacles and staying inside the boundaries.}%
\label{Fig:domain}%
\end{figure}

We build a domain where a particle agent tries to navigate in 2D space with $N$ obstacles to reach the goal position (see illustration in \Cref{Fig:domain}). The system uses double-integrator dynamics.
The reward function is set as $r(s) = -\text{dist}(\text{agent},\ \text{goal})$, where $s$ is the coordination of the particle and $\text{dist}(s_1,\ s_2) = \|s_1 - s_2\|_2$. For obstacle $i$, $\mathcal{X}_i$, the associated cost function is defined as $c_i(s)=2 e^{-\text{dist}(\text{agent},\ \mathcal{X}_{i,c})} + 0.5$ if $s\in \mathcal{X}_i$ and $c_i(s)=0$ otherwise, where $\mathcal{X}_{i,c}$ is the center of $\mathcal{X}_{i}$.

We use the policy gradient reinforcement learning algorithm to solve this problem, where the policy is parameterized by deep neural networks and trained by our LSTM-optimizer. We compare our algorithm against two state-of-the-art safe RL baselines, the Lagrangian~\cite{chow2017risk} and constrained policy optimization (CPO)~\cite{achiam2017constrained} method. 
We use an open-source implementation for these baselines\footnote{\url{https://github.com/openai/safety-starter-agents}}. 
For a reference, we also compare against an unconstrained RL algorithm, proximal policy optimization (PPO), using an open-source implementation\footnote{\url{https://spinningup.openai.com/}}. For reproducibility, the hyperparameters of all the implemented algorithms can be found in the appendix.

\begin{figure}[t]
\centering
\includegraphics[width=0.49\textwidth]{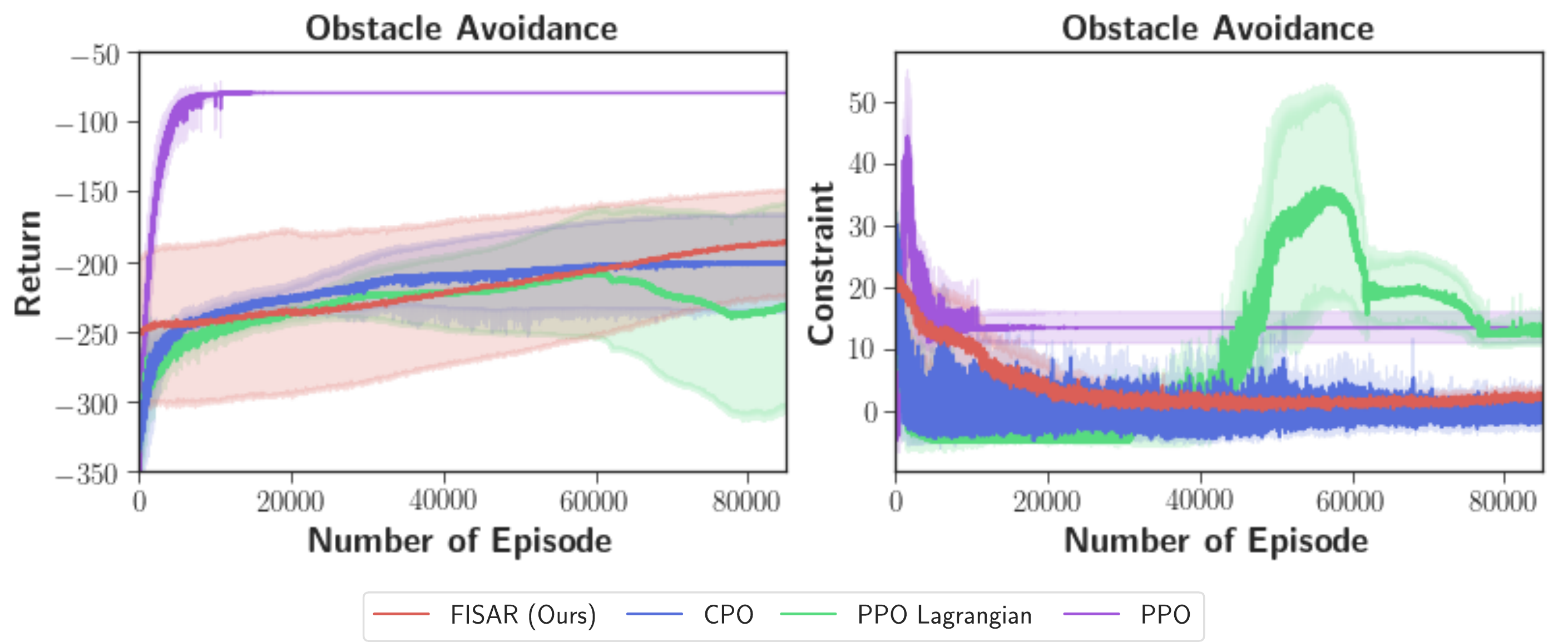}
\label{Fig:constraint}
\vspace{-.6cm}
\caption{
Average performance and $95\%$ confidence interval of the policy over $5$ seeds in the obstacle avoidance domain. FISAR drives the constraint function to decrease satisfaction almost monotonically, but CPO shows much nosier constraint violation and PPO Lagrangian eventually cannot satisfy the constraints. The unconstrained baseline of PPO also violates the constraints as expected.}
\label{Fig:rl}%
\end{figure}

Results of the policy trained by our optimizer and the baselines are demonstrated in \Cref{Fig:rl}. 
There are two notable observations. First, as expected, the unconstrained baseline of PPO achieves the highest return while showing the large constraint violation.
Second, FISAR drives the constraint function to decrease to satisfaction almost monotonically, but CPO's constraint function is much nosier and PPO-Lagrangian eventually cannot satisfy the constraints. 
FISAR achieves the smoother constraint satisfaction with a similar return compared to CPO and PPO Lagrangian baseline.

The failure of the PPO-Lagrangian method may come from the difficulty of solving a minimax problem such that the algorithm gets stuck into a local minimax point, where the primal constraints are not satisfied yet. For the CPO, the successive convexification can be the reason for the oscillation of the constraints within the trust region. However, if the trust region is tuned smaller, the learning process will be slower, resulting in higher sample complexity.

\section{Conclusion}\label{sec:conclusion}
In this paper, we propose to learn a DNN-based optimizer to solve a safe RL problem formulated as CMDP with guaranteed feasibility without solving a constrained optimization problem iteratively. Moreover, the resulting updating dynamics of the variables imply forward-invariance of the safety set. Future work will focus on applying the proposed algorithm in more challenging RL domains as well as more general RL algorithms such as actor-critic and extending it to multiagent RL domains with non-stationarity. The challenge for the latter can partly come from that the safety constraints can be coupled among multiple agents (e.g., collision avoidance), which makes it difficult to get a decentralized policy for each agent.

\section*{Acknowledgements}
Dong-Ki Kim was supported by IBM, Samsung (as part of the MIT-IBM Watson
AI Lab initiative), and Kwanjeong Educational Foundation Fellowship. 
We thank Amazon Web services for computational support.

\appendix
The hyperparameters for our method and the baselines can be found in the following table, where ``NN" and ``lr" stand for ``neural network" and ``learning rate", respectively. For other parameters, we use the default ones in the repositories.
\begin{center}
\begin{tabular}{ c|c||c|c }
\hline\hline
General & \\
 \hline
 Parameter & Value & Parameter & Value\\ 
 \hline
 Policy NN type & MLP & Policy lr & 0.001\\ 
 Policy NN hidden size & 16 & $\gamma$  & 0.99 \\
 Episode length & 100 & & \\
\hline\hline
FISAR (Ours) & \\
 \hline
 Parameter & Value & Parameter & Value\\ 
 \hline
 LSTM hidden number & 128 & $T_{\phi}$ in \eqref{eq:loss} & 120 \\ 
 LSTM hidden layer & 2 & batch size & 24 \\
 LSTM training lr & 0.05 & $\alpha$ in \eqref{eq:lyaponuv} & 20 \\
 $\beta$ in \eqref{eq:optimizer} & 0.001 & &\\ 
 \hline\hline
   CPO, PPO-Lagrangian, PPO &  & \\
 \hline
 Parameter & Value & & \\ 
 \hline
 $\lambda$ (GAE) & 0.95 & & \\
 \hline\hline
\end{tabular}
\end{center}

\bibliographystyle{IEEEtran}
\bibliography{example_paper}

\balance

\end{document}